\documentclass{report}
\usepackage[utf8]{inputenc}
\usepackage{booktabs}
\usepackage{xcolor}
\usepackage{hyperref}
\usepackage{physics}
\usepackage{amsmath} 
\usepackage{amssymb} 
\DeclareMathOperator{\EX}{\mathbb{E}}

\usepackage[most]{tcolorbox}
\usepackage{caption}
\usepackage{subcaption}
\usepackage{textcomp}
\usepackage{dirtytalk}
\newcommand{\mysubfig}[3][width=\linewidth]{%
    \tcbitem\subfloat[#2]{\includegraphics[#1]{#3}}}

\usepackage[pscoord]{eso-pic}
\usepackage{graphics}
\newcommand\BackgroundPic{%
    \put(0,0){%
        \parbox[b][\paperheight]{\paperwidth}{%
            \centering
            \vspace{0.6in}\hspace{5.8in}
            \includegraphics[width=3cm,height=3cm,%
            keepaspectratio]{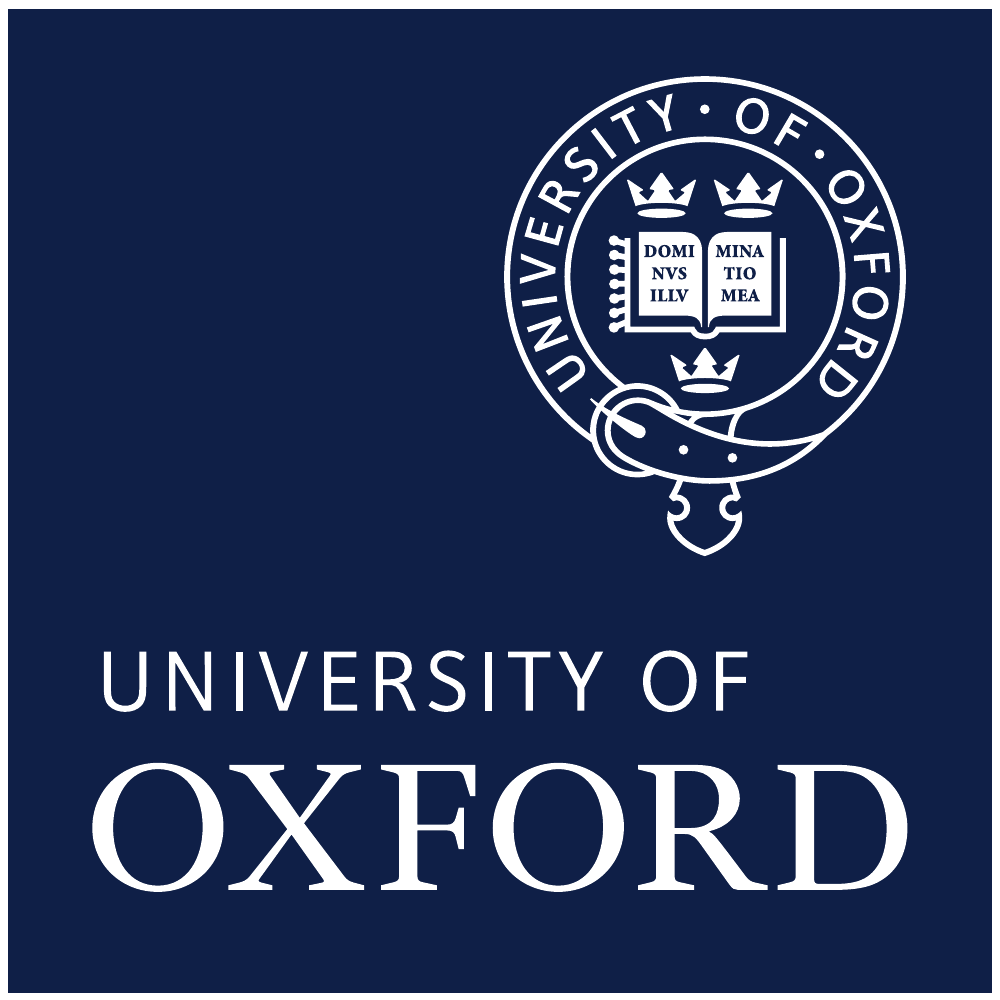}%
            \vfill
}}}

\author{
  M'Charrak, Amine\\
  \texttt{amine.mcharrak@cs.ox.ac.uk}
  \and
  Růžička, Vít\\
  \texttt{vit.ruzicka@cs.ox.ac.uk}
  \and
  Shin, Sangyun\\
  \texttt{sangyun.shin@cs.ox.ac.uk}
  \and
  Vankadari, Madhu\\
  \texttt{madhu.vankadari@cs.ox.ac.uk}
}

\title{\textbf{Tighter Variational Bounds are Not Necessarily Better}\\ Research Report on Implementation, Ablation Study, and Extensions}

\date{April 2021}

\begin{document}
\AddToShipoutPicture*{\BackgroundPic} 

\maketitle

\section{Introduction}

Performing efficient approximate inference and learning for directed probabilistic models is a challenging problem as the latent variables of these models are continuous and have intractable posterior distributions. This problem can be solved by the Variational Bayes approach that involves an optimization step to approximate the intractable posterior. The common mean-field approaches cannot be applied to solve this problem as they demand the analytical solutions of the expected values w.r.t approximate posterior distributions which itself is intractable. In the recent works \cite{kingma2013auto, rezende2014stochastic}, the authors proposed to solve this problem
using a generative model which pairs a top-down generative network with a bottom-up recognition
network. Both networks are jointly trained to maximize a variational lower bound on the intractable model evidence. These models are also known as Variational Autoencoders (VAE).

However, these models assume that the posterior is approximately factorial and that its parameters can be estimated from the input data using nonlinear regression.  \cite{burda2015importance} argues that learning by maximizing a variational lower bound on the log-likelihood encourages the neural network to learn representations where the aforesaid assumptions are satisfied. Despite the benefits of these representations, the learned approximate posterior is not fully expressive because of the imposed constraints. 
Also, they argue that the existing VAE objective harshly penalizes
approximate posterior samples which can not explain the data, even if the recognition network
puts much of its probability mass on good explanations. To address this issue, they propose a generative model named importance weighted autoencoder (IWAE) that shares the VAE architecture but is trained on tighter variational bounds derived from importance weighting. In this method, the recognition network derives many samples (K) from the learned posterior and their weights are averaged with an intuition that as the number of samples K increases, the IWAE lower bound gets closer to true log-likelihood. Also, the importance sampling is expected to help the model to express the posterior that is not generally possible by vanilla VAE formulation. 

In this report, we are going to explain the recent work \cite{rainforth2018tighter} which argues that the aforesaid condition of having tighter evidence lower bounds (ELBO) for improved reconstruction performance is not necessarily better, and doing that deteriorates the overall performance of the model. This work has proven this fact with theoretical and empirical evidence, that increasing the number of importance samples in IWAE \cite{burda2015importance} degrades the signal-to-noise ratio (SNR) of the gradient estimator in the inference network and thereby affecting the full learning process. In other words, even though increasing K decreases the standard deviation of the gradients, it also reduces the magnitude of the true gradient faster, thereby increasing the relative variance of the gradient updates. 
 
Extensive experiments are performed to understand the importance of $K$. These experiments suggest that tighter variational bounds are beneficial for the generative network and looser bounds are preferable for the inference network. With these insights, three methods are introduced namely: the partially importance weighted autoencoder (PIWAE), the multiply importance weighted autoencoder (MIWAE) and the combination importance weighted autoencoder (CIWAE). All these systems entail IWAE as a special case but use the importance weights in different ways to ensure a higher SNR of the gradient estimators. 
In our reproduction report the efficacy of these algorithms is tested on multiple datasets namely, MNIST, and Omniglot. By the end, we demonstrate that the new algorithms are able to approximate the posterior much closer to the true posterior than IWAE and also match the performance of the IWAE generative network or potentially outperforming it in the case of PIWAE.

In this report, we are replicating the original contributions of \cite{rainforth2018tighter} and further add the following contributions:
\begin{enumerate}
    \item CIWAE uses a convex combination ELBO losses from the IWAE and VAE with a constant $\beta$ defining the importance of each of these ELBOs, which generally requires ablation study with various $\beta$ values. We propose to solve this issue by defining the $\beta$ as a learnable parameter with an initial value and let the stochastic optimization to find the best-possible value for the given task.
    
    \item We studied the effect of increasing the number of parameters in the inference and generative network, to understand the effect of model complexity on the performance of the system. 
    
    \item The reproduced paper benchmarked their results with the MNIST dataset only. The authors do not report any of the cross dataset behaviours to understand the generalization aspect of their proposed algorithms. In this report, we perform the evaluation on another dataset named Omniglot along with MNIST. 
    
    \item Finally, we have released our implementations of these methods as an open-source Github repository\footnote{\url{https://github.com/madhubabuv/TightIWAE/}}.
\end{enumerate}

The remainder of this report is organized as follows: The details of the methods proposed in the original paper \cite{rainforth2018tighter} and our contributions are explained in Section~\ref{sec:method}. Various experimental studies with qualitative and quantitative results are delineated in Section \ref{sec:results}. The conclusion remarks and the future scope of this work is presented in Section \ref{sec:conclusion}.

\section{Related Work}
This section details the previous approaches used to learn the deep generative models. Before the raise of VAE, some models were proposed with Botlzman distributions \cite{smolensky1986information}\cite{salakhutdinov2010efficient} with an advantage of modeling tractable and conditional distributions. However, it was impossible to sample data from these learned models or compute the partition function which hampered their progress \cite{salakhutdinov2008quantitative}. On the other hand, a few models are defined using belief networks \cite{neal1992connectionist}\cite{gregor2014deep} which are tractable and easy to sample from but, the conditional distributions become entwined because of the \say{explaining away effect}. Later on, \cite{dayan1995helmholtz} proposed the wake-sleep algorithm using Helmholtz machines which approximates the posterior by learning a recognition network along with a generative network. Unfortunately, each network needs a different objective function during the training. Deep auto-regressive networks \cite{gregor2014deep} solved this problem of two objectives by training their network using a single variational lower bound. In this similar line of research, \cite{mnih2014neural} proposed another algorithm that reduces the stochasticity in the gradient updates using a third network to predict reward baselines (REINFORCE algorithm \cite{williams1992simple}) that minimizes the expected square of the learning signal during the training. 

Variational autoencoders~\cite{kingma2015variational}\cite{rezende2014stochastic}, as explained in the introduction section, are also generative models and also trained using the same objective functions same as \cite{mnih2014neural}. Instead of using REINFORCE to reduce the variance in the gradient updates, they use a technique named ``backprop through a random number generator" \cite{williams1992simple} for the reparametrization of the random choices. In VAEs, the modes are described with a simple distribution (Gaussian) followed by a deterministic mapping instead of a sequence of random choices. 

There are also works by \cite{neal1990learning} and \cite{ba2014multiple} that tried deriving log-probability lower bounds by avoiding recognition networks and perform the inference using importance sampling from the prior. In another work \cite{gogate2007studies}, a different variety of graphical inference models is proposed using importance weighing. Similar to IWAE \cite{burda2015importance}, another model named Reweighted Wake-sleep \cite{bornschein2014reweighted} is another generative model that extends the original Wake-sleep algorithm with a recognition network. All these networks implicitly assume the tightening the variational bounds is good for both generative and recognition networks which is proven not to be necessary for \cite{rainforth2018tighter}.

\section{Method}
\label{sec:method}

In order to explain the authors three newly proposed algorithms (MIWAE, PIWAE, and CIWAE) we first describe the basic concepts of the VAE methodology and its optimization objective, the ELBO. 
Moreover, we highlight how the IWAE, an extension of the VAE, conceptually differs from the VAE modeling objective. 
Finally, we provide technical insights into how each of the three proposed estimators address the diminishing SNR modeling issue of IWAEs. 

\subsection{VAE and evidence lower bound (ELBO)}
The Variational Autoencoder (VAE) is a probabilistic deep learning model for density estimation and representation learning of continuous latent variable models $p_{\theta}(x \vert z)$ with intractable posterior distributions $p_{\theta}(z \vert x)$ over latent variables $z$ given data observations $x$ \cite{nowozin2018debiasing}.
The distribution of interest arises by marginalization over the latent variables $z$
\begin{equation}\label{eqn:likelihood}
p(x) = \int p_{\theta}(x \vert z) p_{\theta}(z) \mathrm{d}z
\end{equation}

with $p_{\theta}(z)$ being the prior distribution and often times set to be standard Normal distribution $\mathcal{N}(0,I)$.
This marginal probability distribution (or likelihood) $p(x)$ is intractable when the latent variable space $\mathcal{Z}$ is high dimensional due to the expensive marginalization operation. 
As such, learning the parameters $\theta$ via maximum likelihood estimation (MLE) becomes infeasible.\\

Therefore, in order to avoid the exponential marginalization costs in Equation (\ref{eqn:likelihood}) the authors of \cite{kingma2013auto} propose to instead optimize for the evidence lower-bound (ELBO) by introducing an inference network $q_{\phi}(z \vert x)$ with learnable parameters $\phi$. 
This enables us to cast the inference problem as an optimization problem by learning an analytic approximation $q_{\phi}(z \vert x)$ of the true posterior distribution $p_{\theta}(z \vert x)$ which is intractable because of its dependence on the difficult to compute marginal likelihood
\begin{equation}
    p_{\theta}(z \vert x) = \frac{p_{\theta}(x,z)}{p_{\theta}(x)} \mathrm{.}
\end{equation}

Given the fact, that the logarithm is a monotonically increasing function of its argument, one can instead infer the parameters $\theta$ by maximizing the log likelihood, $\mathrm{log} \: p(x)$, instead of the likelihood.
We can rewrite Equation (\ref{eqn:likelihood}) by applying the logarithm of the likelihood. To make the integral with respect to $z$ tractable, we can use the approximate posterior $q_{\phi}(z \vert x)$ instead of the intractable true posterior $p_{\theta}(z \vert x)$.
With these adjustments we can now derive the tractable VAE evidence lower bound ($ELBO_{VAE}$) to the log likelihood as follows

\begin{eqnarray}
    \mathrm{log} \: p(x) &=& \mathrm{log} \: \int p_{\theta}(x \vert z) p_{\theta}(z) \mathrm{d}z \nonumber\\
    &=& \mathrm{log} \: \int \frac{p_{\theta}(x \vert z) p_{\theta}(z)}{q_{\phi}(z \vert x)} \; q_{\phi}(z \vert x) \mathrm{d}z \nonumber\\
    &=& \mathrm{log} \EX_{q_{\phi}(z \vert x)} \left[ \frac{p_{\theta}(x \vert z) p_{\theta}(z)}{q_{\phi}(z \vert x)} \right] \label{eqn:ELBO_VAE}\\
    &\geq& \EX_{q_{\phi}(z \vert x)} \left[ \mathrm{log} \: \frac{p_{\theta}(x \vert z) p_{\theta}(z)}{q_{\phi}(z \vert x)} \right] =: \mathcal{L}_{VAE} (= ELBO_{VAE}) \mathrm{.} \nonumber
\end{eqnarray}

The inference problem is now an optimization problem because one needs to find a variational distribution $q_{\phi}(z \vert x)$ that maximizes $\mathcal{L}_{VAE}$.
To make the computation tractable and the inference network expressive enough of the input data $x$, the approximate posterior is chosen to be factorial (e.g., a Gaussian distribution with diagonal covariance matrix) and its parameters $\phi$ are determined by a non-linear function of the data (e.g., a neural network).
The gap between the evidence lower bound $\mathcal{L}_{VAE}$ and the log likelihood $\mathrm{log} \: p(x)$ can be quantified by reformulating Equation \ref{eqn:ELBO_VAE} such that

\begin{eqnarray}
    \mathcal{L}_{VAE} &=& \EX_{q_{\phi}(z \vert x)} \left[ \mathrm{log} \: \frac{p_{\theta}(x \vert z) p_{\theta}(z)}{q_{\phi}(z \vert x)} \right] 
    = \EX_{q_{\phi}(z \vert x)} \left[ \mathrm{log} \: \frac{p_{\theta}(z \vert x) p_{\theta}(x)}{q_{\phi}(z \vert x)} \right] \nonumber\\
    &=& \EX_{q_{\phi}(z \vert x)} \left[ \mathrm{log} \: p_{\theta}(x) \right] + \EX_{q_{\phi}(z \vert x)} \left[ \mathrm{log} \: \frac{p_{\theta}(z \vert x)}{q_{\phi}(z \vert x)} \right] \nonumber\\
    &=& \mathrm{log} \: p_{\theta}(x) + \EX_{q_{\phi}(z \vert x)} \left[ \mathrm{log} \: p_{\theta}(z \vert x) - \mathrm{log} \: q_{\phi}(z \vert x) \right] \nonumber\\
    &=& \mathrm{log} \: p_{\theta}(x) - \EX_{q_{\phi}(z \vert x)} \left[ \mathrm{log} \: q_{\phi}(z \vert x) - \mathrm{log} \: p_{\theta}(z \vert x) \right] \nonumber\\
    &=& \mathrm{log} \: p_{\theta}(x) - \EX_{q_{\phi}(z \vert x)} \left[ \mathrm{log} \: \frac{q_{\phi}(z \vert x)}{p_{\theta}(z \vert x)} \right] \nonumber\\
    &=& \mathrm{log} \: p_{\theta}(x) - D_{KL}\left(q_{\phi}(z \vert x) \, \vert \vert \, p_{\theta}(z \vert x)\right) \label{eqn:KLdivergence}
\end{eqnarray}

where we use the definition of the Kullback-Leibler (KL) divergence
\begin{equation*}
    D_{KL}(Q \, \vert \vert \, P) = \EX_{Q} \left[ \mathrm{log} \left( \frac{Q}{P} \right) \right]
\end{equation*}
from the distribution $P$ to the distribution $Q$ for the last equivalence.
From Equation \ref{eqn:KLdivergence} we observe that the difference between the marginal log likelihood $\mathrm{log} \: p_{\theta}(x)$ and the ELBO ($\mathcal{L}_{VAE}$) is the KL divergence of the approximate posterior distribution $p_{\phi}(z \vert x)$ and the true posterior distribution $p_{\theta}(z \vert x)$.
Given the non-negative KL divergence and the log likelihood of the data being a constant, one can maximize the ELBO by minimizing the KL divergence term between these two distributions.
When the learned approximate posterior distribution of the inference network is identical to the true posterior distribution of the true latent variable model, then the KL divergence is zero and the ELBO equals the log likelihood of the data.\\

In practice a simple Monte Carlo estimator of the ELBO $\mathcal{L}_{VAE}$ can be used to approximate the expectation in Equation (\ref{eqn:ELBO_VAE}) by the sample mean such that

\begin{equation} \label{eqn:MC_estimate_ELBO}
    \hat{\mathcal{L}}_{VAE} = \frac{1}{L} \sum_{i = 1}^{L} \mathrm{log} \: \frac{p_{\theta}(x \vert z_i) p_{\theta}(z_i)}{q_{\phi}(z_i \vert x)} \mathrm{.}
\end{equation}

This estimator is unbiased because $\EX_{z_i \sim q_{\phi}{z \vert x}} \left[ \mathcal{L}_{VAE} \right] = \mathcal{L}$.
Concurrent maximization of Equation (\ref{eqn:ELBO_VAE}) with respect to both neural networks' parameters ${\theta, \phi}$ results in approximate maximum likelihood estimation of $p(x)$. 
This can be achieved through end-to-end learning using back propagation \cite{backprop1986} and the reparameterisation trick \cite{kingma2015variational} to allow for gradient optimisation methods despite the stochastic operations performed for retrieving samples $z_i$ from the learned approximate posterior distribution over latent variables $q_{\phi}(z \vert x)$.\\

From an architectural perspective, the VAE fits an encoder, called inference network and typically parameterised as a neural network, to learn an arbitrary complex mapping between a data point $x$ from the data space $\mathcal{X}$ and the latent variable $z$ from the unobservable latent space $\mathcal{Z}$ with the distribution over unobserved variables $p_{\theta}(z \vert x)$.
Using samples from the posterior distribution, a decoder, called generative network and also parameterised by a neural network, learns a mapping from the unobservable latent space $\mathcal{Z}$ back into the input space of our data observations $\mathcal{X}$. 
As such, the generative network enables the fitting of a conditional distribution $p_{\theta}(x \vert z)$.
This distribution describes the generating process of the dataset $\mathcal{D} = \{x_i\}_{i=1}^{N}$, with each $x_i$ being sampled i.i.d from the true generative process $p_{\theta}(x \vert z)$.

\subsection{IWAE: A multiple sample VAE extension with tighter bounds}

The VAE objective is limited in its ability to infer complex posterior distributions due to the strong assumptions it makes, e.g., a factorial true posterior distribution that can be learned with a feed-forward neural network \cite{burda2015importance}.
For most real-world datasets these assumptions are oftentimes not met and hence the VAE objective results in learning an oversimplified approximate factorial posterior distribution.
To alleviate these restrictions, \cite{burda2015importance} propose a tighter evidence lower bound, the IWAE ELBO, which employs a multiple (i.e., $K$) sample importance-sampling strategy on the marginal likelihood.\\

The IWAE objective can be derived by approximating the expectation inside the logarithm of Equation (\ref{eqn:ELBO_VAE}) before applying Jensen's inequality to retrieve the IWAE evidence lower bound.
The authors employ Monte Carlo to estimate the expression inside the logarithm using the sample mean based on $K$ samples from the approximate posterior distribution $q_{\phi}(z \vert x)$.
Using the result from Equation (\ref{eqn:ELBO_VAE}), we can denote the $ELBO_{IWAE}$ as

\begin{eqnarray}
    \mathrm{log} \: p(x) &=& \mathrm{log} \EX_{q_{\phi}(z \vert x)} \left[ \frac{p_{\theta}(x \vert z) p_{\theta}(z)}{q_{\phi}(z \vert x)} \right] \nonumber\\
    &=& \mathrm{log} \left( \EX_{z_1, \dots, z_K \sim q_{\phi}(z \vert x)} \left[ \frac{1}{K} \sum_{i=1}^{K} \frac{p_{\theta}(x \vert z_i) p_{\theta}(z_i)}{q_{\phi}(z_i \vert x)} \right] \right) \label{eqn:ELBO_IWAE}\\
    &\geq& \EX_{z_1, \dots, z_K \sim q_{\phi}(z \vert x)} \left[ \mathrm{log} \left( \frac{1}{K} \sum_{i=1}^{K} \frac{p_{\theta}(x \vert z_i) p_{\theta}(z_i)}{q_{\phi}(z_i \vert x)} \right) \right] =: \mathcal{L}_{IWAE}^{K} (= ELBO_{IWAE}) \mathrm{.} \nonumber
\end{eqnarray}

If we choose to optimize the IWAE ELBO in Equation (\ref{eqn:ELBO_IWAE}) with $K=1$, the VAE and IWAE ELBO estimators are equivalent.
As $K$ increases we get that $\mathcal{L}_{IWAE}^{K} \xrightarrow{} \mathrm{log} \: p(x)$.
Finally, for any $K \geq 1$ it can be shown that the importance-weighted autoencoder defines a variational inference strategy that allows for arbitrarily tight evidence lower bounds \cite{cremer2017reinterpreting} such that

\begin{equation*}
    \mathrm{log} \: p(x) \geq \mathcal{L}_{IWAE}^{K+1} \geq \mathcal{L}_{IWAE}^{K} \geq
    \mathcal{L}_{IWAE}^{K=1} = \mathcal{L}_{VAE} \mathrm{.}
\end{equation*}

The benefits of a tighter evidence lower bound for the IWAE comes with the cost of getting a biased estimator as compared to the VAE ELBO which enabled unbiased estimates of the marginal log likelihood \cite{nowozin2018debiasing}.
Compared to the VAE with $K=1$, setting $K>1$ results in significant improvements for learning the generative model as measured by the marginal log likelihood of the data, $\mathrm{log} \: p(x)$.\\

Using Equation (\ref{eqn:ELBO_IWAE}) the gradient estimator of the IWAE ELBO for the inference and generative network has the form

\begin{equation} \label{eqn:IWAE_gradient}
    \Delta_{K} = \nabla_{\theta, \phi} \: \mathrm{log} \frac{1}{K} \sum_{k=1}^{K} w_k
\end{equation}

where $w_k = \frac{p_{\theta}(x \vert z_k) p_{\theta}(z_k)}{q_{\phi}(z_k \vert x)}$ and $z_k \stackrel{i.i.d}{\sim} q_{\phi}(z_k \vert x)$.
\subsection{Problem of tighter variational bounds}
It was believed that increasing K ensures tighter variational bounds, leading to the better performance of VAE. However, the paper we chose \cite{rainforth2018tighter} questions this, showing that increasing K doesn't always lead to better performance. To prove this, signal-to-noise ratio (SNR) is introduced for estimating the relative accuracy of gradient. SNR is defined as follows:

\begin{equation}
    SNR_{M,K}(\theta)=|\EX[\Delta_{M,K}(\theta)]/\sigma_{M,K}[(\theta)]|
\end{equation}
where $\sigma$ stands for the standard deviation of a random variable.
The intuition behind SNR is that, although high SNR does not always mean that the gradient is well calculated without noise, low SNR is always detrimental due to the dominant noise in the estimates. 
To formally investigate this, \cite{rainforth2018tighter} derives the convergence rates of $SNR_{\phi}$ and $SNR_{\theta}$ as follows:
\begin{equation}
    SNR_{M,K}(\theta)=\sqrt{M}\abs{\frac{\sqrt{K}\nabla_{\theta}Z-\frac{1}{2Z\sqrt{K}}\nabla_{\theta}\frac{Var[w_{1,1}]}{Z^{2}}+O(\frac{1}{K^{3/2}})}{\sqrt{\EX[w^{2}_{1,1}(\nabla_{\theta}log w_{1,1}-\nabla_{theta}log Z)^2]}+ O(\frac{1}{K})}}
\label{eqn:SNR_inference}
\end{equation}
\begin{equation}
    SNR_{M,K}(\phi) = \sqrt{M}\abs{\frac{\nabla_{\phi}Var[w_{1,1}]+O(\frac{1}{K})}{2Z\sqrt{K}\sigma[\nabla_{\phi}w_{1,1}]+O(\frac{1}{\sqrt{K}})}}
\end{equation}
where $Z=p_{\theta}(x)$ is the true marginal likelihood.

Intuitively, according to $SNR_{M,K}(\theta)$ and $SNR_{M,K}(\phi)$, it is clear that increasing $M$ always leads to higher $SNR_{M,K}(\theta)$ and $SNR_{M,K}(\phi)$, benefiting both inference and generative networks. 
However, when increasing $K$, $SNR_{\theta}$ becomes larger while $SNR_{\phi}$ becomes lower, suggesting that tighter variational bounds by increasing $K$ is only good for the generative network and can be detrimental for the inference network. Based on this insight, \cite{rainforth2018tighter} proposes three improved methods addressing the problem of the tighter variational bounds.

\subsection{Multiply importance weighted auto-encoder (MIWAE)}

From Equations \ref{eqn:ELBO_IWAE} and \ref{eqn:IWAE_gradient} we observe that we are using only a single sample of $\mathrm{log} \frac{1}{K} \sum_{k=1}^{K} w_k$ to estimate importance weighted ELBO and its gradients with respect to the parameters of the inference and generative network.

As such, the first natural approach would be to reduce the variance of the gradient estimates $\nabla_{\theta, \phi} \mathcal{L}_{IWAE}^{K}$ by taking multiple samples, e.g., $M$, to estimate the gradients of the ELBO.
As stated in \cite{rainforth2018tighter}, $M$ does not affect the true value of the gradients of the lower bound.
However, $K$, the number of samples used for the IWAE, does affect the bound by making it tighter.\\

The first of the three newly proposed algorithms proposed to address the issue of a vanishing SNR in the inference network is called MIWAE. 
This algorithm uses $M$ instead of 1 sample in order to estimate $\mathcal{L}_{IWAE}^{K}$ and its gradients. 
The gradient estimates for the multiply importance weighted autoencoder are computed as follows:

\begin{equation}
    \Delta_{M, K} = \frac{1}{M} \sum_{m=1}^{M} \left[ \nabla_{\theta, \phi} \: \mathrm{log} \: \left( \frac{1}{K} \sum_{k=1}^{K} w_{m,k} \right) \right]
\end{equation}

with $w_{m,k} = \frac{p_{\theta}(x \vert z_{m,k}) p_{\theta}(z_k)}{q_{\phi}(z_{m,k} \vert x)} = \frac{p_{\theta}(z_{m,k}, x) p_{\theta}(z_k)}{q_{\phi}(z_{m,k} \vert x)} $ and $z_{m,k} \stackrel{i.i.d}{\sim} q_{\phi}(z_{m,k} \vert x)$.\\

Assuming a fixed computation budget with $T = MK$ Monte Carlo samples, we can retrieve two special cases of the MIWAE namely:

\begin{itemize}
    \item for $K=1$ and $M=T$ we get a Monte Carlo estimator of the VAE lower bound objective as shown in Equation (\ref{eqn:MC_estimate_ELBO})
    \item for $K=T$ and $M=1$ we get the corresponding Monte Carlo estimate for the IWAE lower bound objective.
\end{itemize}

The motivation for this approach is based on the observation that the SNR of the IWAE inference network increases with $O(\sqrt{M/K})$.
As such, it should be possible the mitigate the detrimental effect of larger values of $K$ by also increasing the number of samples, $M$, used for Monte Carlo estimation of the evidence lower bound in the IWAE.
To be able to fairly compare the different algorithms proposed, the overall computation budget, $T$, will be held fixed for all proposed algorithms and evidence lower bounds.

\subsection{Combination importance weighted auto-encoder (CIWAE}

The second proposed algorithm, CIWAE, employs a convex combination of the variational lower bounds for the VAE and IWAE objective respectively. 
We have derived and discussed both evidence lower bounds, $\mathcal{L}_{VAE}$ and $\mathcal{L}_{IWAE}^{K}$, in the previous subsections.
The evidence lower bound of the CIWAE takes the form

\begin{equation}\label{eqn:CIWAE_ELBO}
    \mathcal{L}_{CIWAE}^{K} = \beta \mathcal{L}_{VAE} + (1- \beta) \mathcal{L}_{IWAE}^{K} 
\end{equation}

with $\beta \in \left[0,1\right]$ being a combination parameter that acts as a weighting on each of ELBO terms.
Given that $\mathcal{L}_{CIWAE}^{K}$ is a combination of a the tighter IWAE bound and the looser VAE bound, it is trivial to see that

\begin{equation*}
    \mathcal{L}_{IWAE}^{K} \geq \mathcal{L}_{CIWAE}^{K} \geq \mathcal{L}_{VAE} \mathrm{.}
\end{equation*}

Finally, the gradient estimator of the CIWAE algorithm takes on the following form

\begin{equation} \label{eqn:gradient_estimate_CIWAE}
    \Delta_{K, \beta}^{CIWAE} = \nabla_{\theta, \phi}\left( \beta \frac{1}{K} \sum_{k=1}^{K} \mathrm{log} \: w_k + (1-\beta) \mathrm{log} \: \left( \frac{1}{K} \sum_{k=1}^{K} w_k \right) \right) \mathrm{.}
\end{equation}

The idea behind proposing the convex combination of two different ELBOs comes from the observations made for the SNR of the inference network.
When the gradient estimates of the IWAE vanish, the non-zero gradient estimates of the VAE (even for small $\beta$) should enable the CIWAE modeling objective to alleviate SNR issues in the inference network (i.e., the gradient estimates with respect to $\phi$). 
For the CIWAE algorithm, the SNR increases with $O(\sqrt{MK})$.

\subsection{Partially importance weighted auto-encoder (PIWAE)}

Finally, the last proposed algorithm is based on the insight, that an increased $K$ in the IWAE objective is good for the generative network, however, detrimental to the learning signal in the inference network.

Therefore, the authors suggest to use different evidence lower bound targets for the inference network, $q_{\phi}(z \vert x)$, and the generative network, $p_{\theta}(x \vert z)$.
To achieve this, the third and last proposed method, called partially IWAE, PIWAE, is introduced.
The strategy of this variational inference method is to optimize the inference network based on the MIWAE target, allowing SNR adjustments by increasing the sample size $M$, and the generative network on the IWAE target, in order to benefit from the strictly tighter bound as $K$ increases and thus leading to improved empirical gains in terms of the generative model's marginal likelihood estimation.

The corresponding PIWAE gradient estimates of the ELBO with respect to the inference network's parameters, $\phi$, and the generative network's parameters, $\theta$ are given by

\begin{equation} \label{eqn:gradient_estimate_PIWAE_inference}
    \Delta_{M, L}^{PIWAE} (\phi) = \frac{1}{M} \sum_{m=1}^{M} \left[ \nabla_{\phi} \: \mathrm{log} \: \left( \frac{1}{L} \sum_{l=1}^{L} w_{m,l} \right) \right]
\end{equation}

\begin{equation} \label{eqn:gradient_estimate_PIWAE_generative}
        \Delta_{K}^{PIWAE} (\theta) = \nabla_{\theta} \: \mathrm{log} \frac{1}{K} \sum_{k=1}^{K} w_k
\end{equation}

where $L$ is the number of importance-weighted samples used for the evidence lower bound with regards to the optimization for parameters of the inference network.
Instead of simultaneous optimization using a single target lower bound as in all other VAE and IWAE algorithms above, we now now have to optimize for two separate ELBO targets.
This is achieved by using two separate optimizers, one which only modifies the parameters, $\phi$ of the inference network and another optimizer, which only affects the parameters, $\theta$, of the generative network.
In our implementation, we first update the encoding inference network and then the decoding generative network.

\subsection{$\beta$ as Learnable Parameter in CIWAE}
Three proposed methods, such as MIWAE, CIWAE and PIWAE, are effectively alleviating the issue of SNR by different means. One of the possible improvements we found was to reduce the number of hyper-parameter that require cumbersome tuning. For instance, as was addressed in the paper, setting the \say{relatively low} $\beta$ in CIWAE would help when the gradient of IWAE becomes very small. However, defining a \say{relatively small} value for the different dataset might not be a trivial task, requiring a large number of experiments. Also, in domains where the gradient of IWAE is always big enough, setting the small $\beta$ would lead to suboptimal performance of the network. Therefore, it is natural to question if making the $\beta$ learnable would be advantageous for VAE. 

We use gradient descent for optimizing $\beta$ since its optimization can be simultaneously performed with the optimization of VAE. Note that, however, optimizing $\beta$ can be performed by any other optimization techniques.
By defining $ELBO_{VAE}$ as $ELBO$ for VAE and $ELBO_{IWAE}$ as $ELBO$ using IWAE with $K$, the objective for optimizing $\beta$ is:
\begin{equation}
    \min_{\beta} \beta ELBO_{VAE} + (1-\beta)ELBO_{IWAE}
\end{equation}
\section{Results}
\label{sec:results}

\subsection{Reproducing the result on MNIST and Omniglot}
To reproduce the results of the original paper \cite{rainforth2018tighter} we chose two benchmark datasets, the MNIST dataset of handwritten digits and the Omniglot dataset of diverse handwritten characters. We follow the dataset processing steps described in \cite{burda2015importance}, and use the binarized versions of the two datasets alonside the official train and test data splits. Similarly, we chose the same model architectures as those described in \cite{burda2015importance}, namely the model described as having a single stochastic layer. This model has two linear layers with a hidden dimension of 200 in both the Encoder and Decoder networks, and latent dimension of 50.  As an exploration of these parameters, we chose to also train a larger model in terms of the number of parameters with the same layer layout and hidden dimension of 400 and latent dimension of 20 (note that these are also the basic parameters used in the official Pytorch example\footnote{\url{https://github.com/pytorch/examples/blob/master/vae/main.py}}). The number of trainable parameters in this proposed larger model is more than double when compared with the referential model (smaller model has 425284 and the larger model has 973624 parameters).

\begin{figure}[tb]
\begin{tcbitemize}[
    raster columns=2,
    raster halign=center,
    raster every box/.style={blankest},
    fonttitle=\small\bfseries 
    ]
\mysubfig{IWAE$_{64}$}{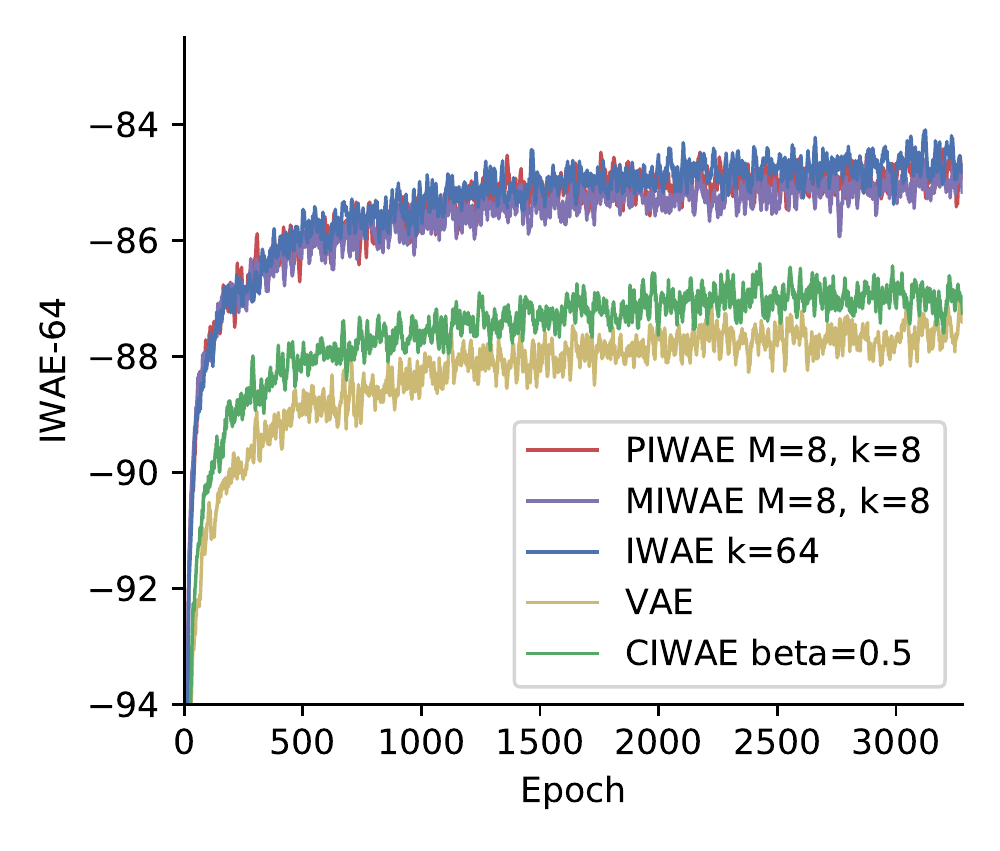}
\mysubfig{log\^{p}(x)}{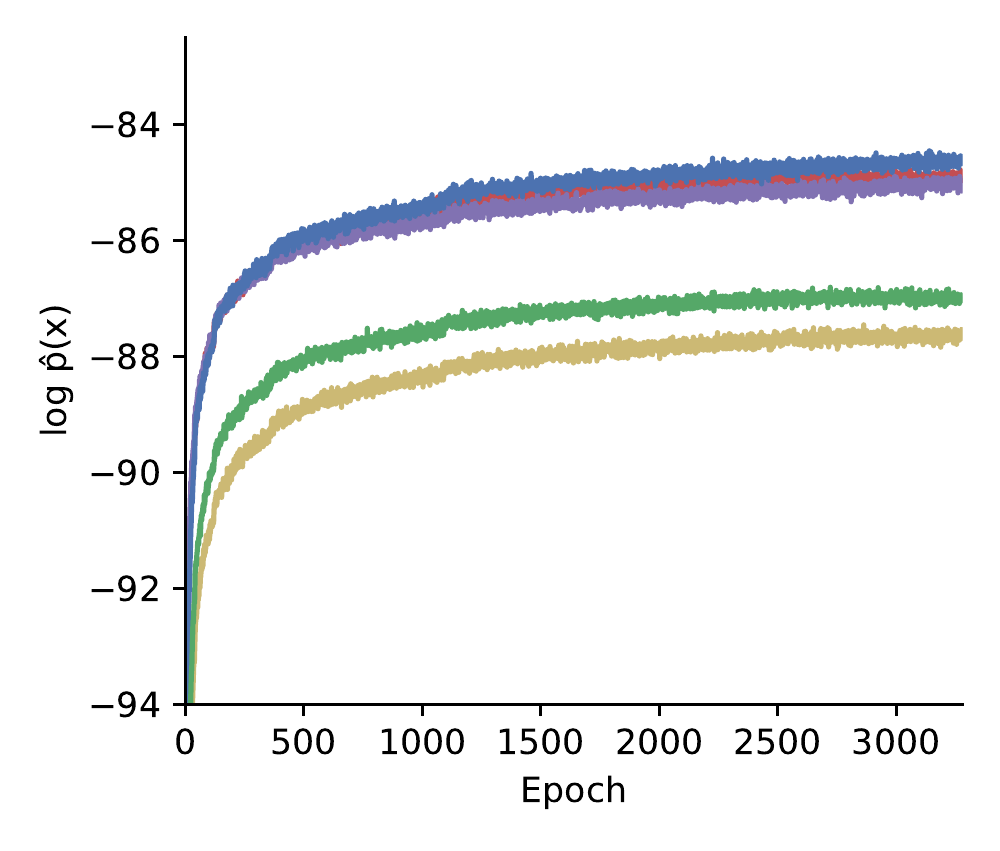}

\end{tcbitemize}

\caption{The metrics of IWAE$_{64}$ and log\^{p}(x) of referential models trained on the MNIST dataset across the training progress. We used rolling window of 10 for better clarity of the IWAE$_{64}$ plot. Note that the figures are shared between (a) and (b).} \label{fig:reproduced_originalmodel}
\end{figure}

We have used the same experiment conditions when reproducing the results of the original paper \cite{rainforth2018tighter}. Namely we train these models for 3280 epochs, use the Adam optimizer (with default parameters) and use the learning rate scheduling described in \cite{burda2015importance}. The final reproduction results are reported in Table \ref{fig:reproduced_table} and the training progress can be in detail explored in Figure \ref{fig:reproduced_originalmodel}.

\begin{figure}[tb]
\begin{tcbitemize}[
    raster columns=2,
    raster halign=center,
    raster every box/.style={blankest},
    fonttitle=\small\bfseries 
    ]
\mysubfig{IWAE$_{64}$}{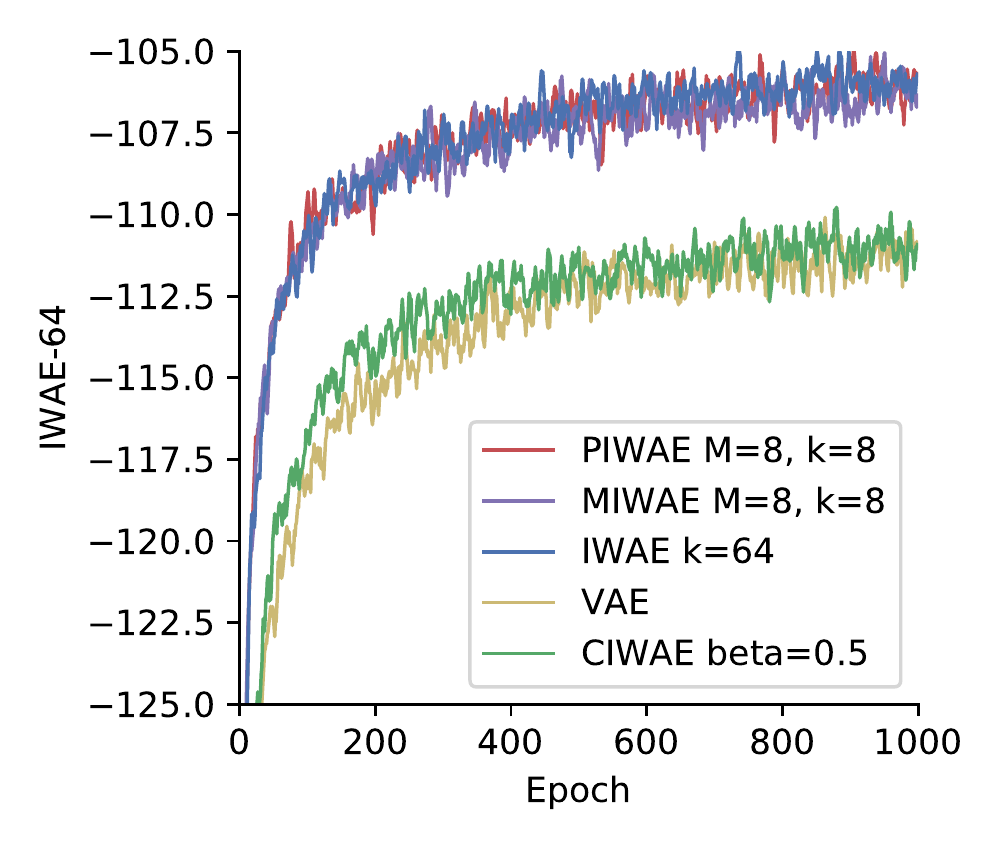}
\mysubfig{log\^{p}(x)}{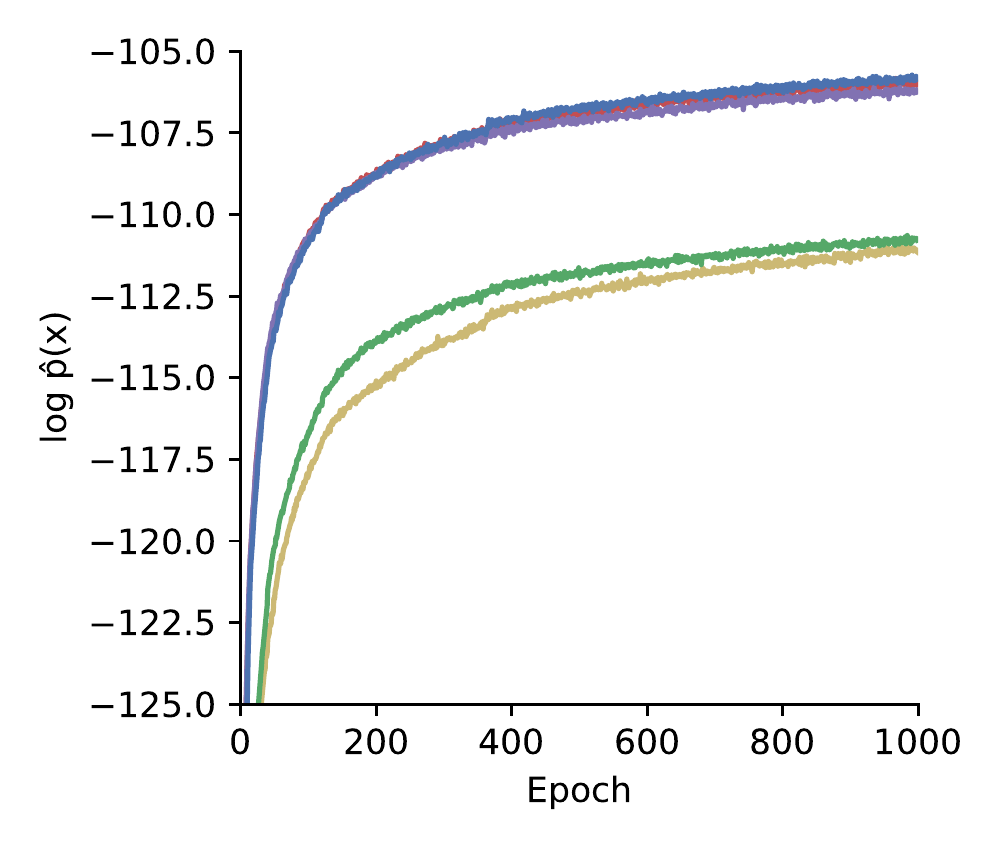}

\end{tcbitemize}

\caption{The metrics of IWAE$_{64}$ and log\^{p}(x) of referential models trained on the Omniglot dataset across the training progress. We used rolling window of 5 in (a). Note that we have trained these models for 1000 epochs.} \label{fig:new_omniglot}
\end{figure}

\begin{figure}[tb]
\begin{tcbitemize}[
    raster columns=2,
    raster halign=center,
    raster every box/.style={blankest},
    fonttitle=\small\bfseries 
    ]
\mysubfig{IWAE$_{64}$}{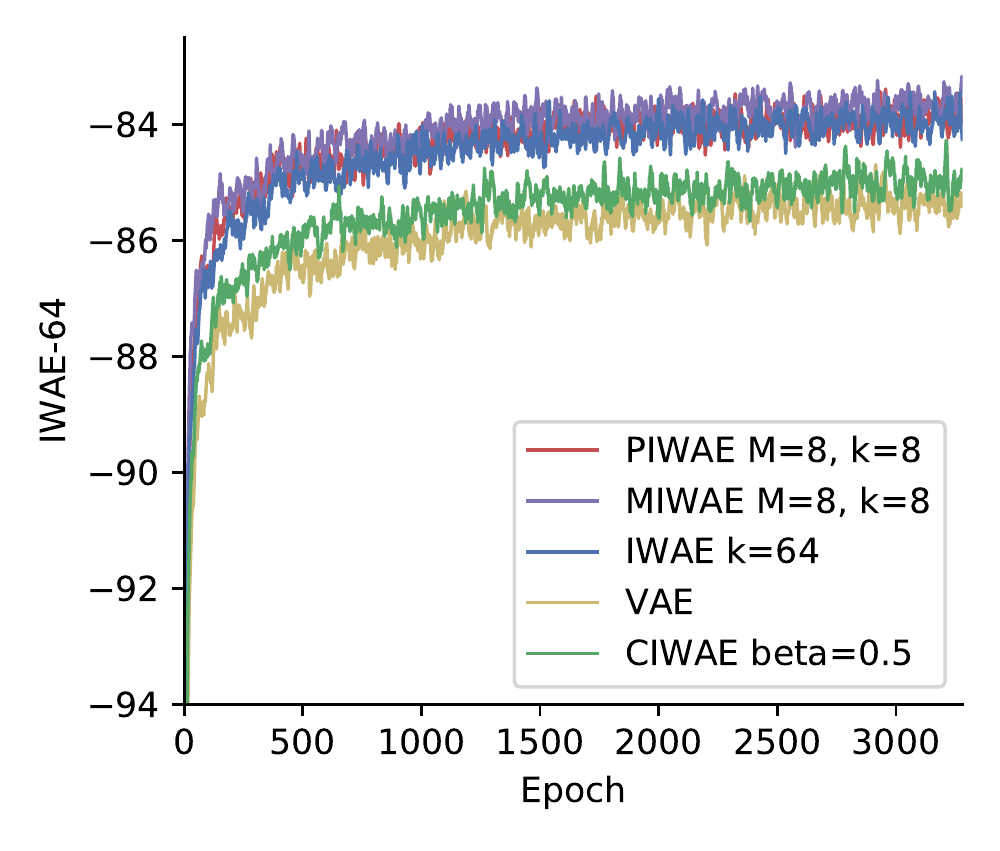}
\mysubfig{log\^{p}(x)}{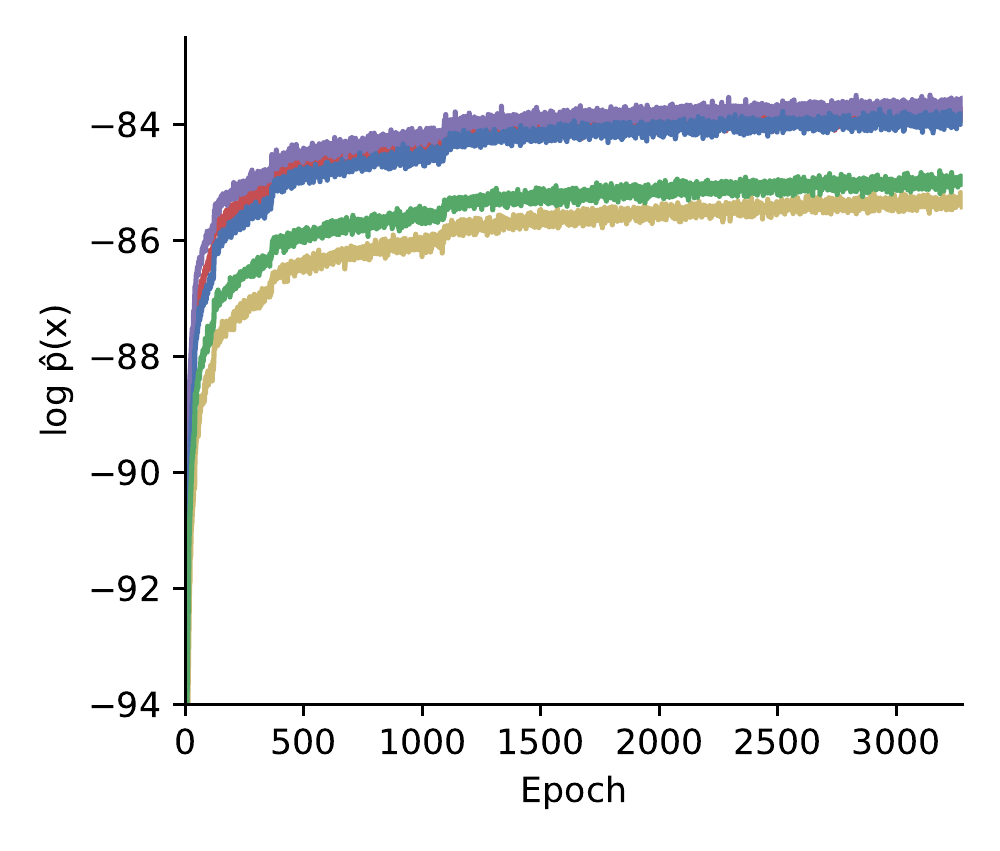}

\end{tcbitemize}

\caption{The metrics of IWAE$_{64}$ and log\^{p}(x) on a larger version of models. We used rolling window of 10 in (a).} \label{fig:new_bigger}
\end{figure}

When evaluating the tested models on the Omniglot dataset, we lowered the number of epochs to 1000 due to limited available compute and differentiated learning plots as is shown in Figure \ref{fig:new_omniglot}. Exploratory results of the larger version of the model trained on MNIST can be seen in Figure \ref{fig:new_bigger}.

\begin{table}[]
\caption{Referential and tested larger model trained for 3280 epochs on the MNIST dataset. Numbers in the brackets indicate the $(M, K)$ and the $\beta$.}
\label{fig:reproduced_table}
\begin{tabular}{|c|ccccc|}
\hline
Metric    & IWAE   & PIWAE$_{(8, 8)}$ & MIWAE$_{(8, 8)}$ & CIWAE$_{\beta = 0.5}$ & VAE    \\ \hline
IWAE-64   & -84.64 & -84.72       & -84.98       & -87.05           & -87.26 \\ 
log\^{p}(x)  & -84.64 & -84.90       & -85.04       & -87.00           & -87.66 \\ 
-KL(Q$||$P) & 0.00  & 0.19        & 0.06        & -0.05            & 0.40   \\ \hline \hline
Larger model    & IWAE   & PIWAE$_{(8, 8)}$ & MIWAE$_{(8, 8)}$ & CIWAE$_{\beta = 0.5}$ & VAE    \\ \hline
IWAE-64   & -83.85                   & -83.86                           & -83.47                           & -84.92                               & -85.33                  \\
log\^{p}(x))  & -83.92                   & -83.84                           & -83.66                           & -84.98                               & -85.31                  \\
-KL(Q$||$P) & 0.06                     & -0.02                            & 0.19                             & 0.06                                 & -0.02         \\ \hline 
\end{tabular}
\end{table}

\begin{figure}[tb]
\begin{tcbitemize}[
    raster columns=6,
    raster halign=center,
    raster every box/.style={blankest},
    fonttitle=\small\bfseries 
    ]
\mysubfig{original \\ input}{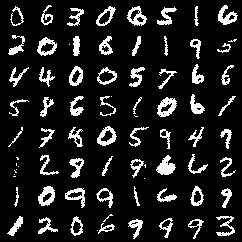}
\mysubfig{IWAE \\ 0.816$\pm$0.051}{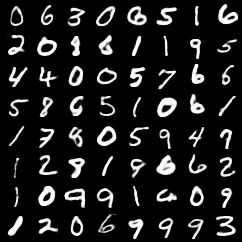}
\mysubfig{PIWAE \\ 0.832$\pm$0.047}{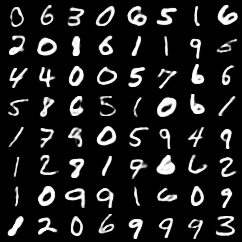}
\mysubfig{MIWAE \\ \textbf{0.833$\pm$0.048}}{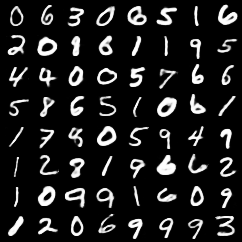}
\mysubfig{CIWAE \\ \textbf{0.836$\pm$0.049}}{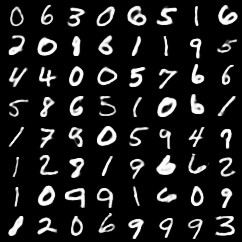}
\mysubfig{VAE \\ 0.830$\pm$0.052}{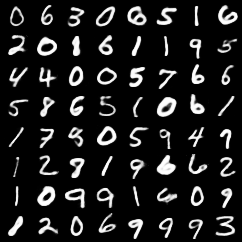}
\end{tcbitemize}

\caption{Qualitative and quantitative analysis of the reconstructed images on the MNIST dataset with the referential model size. We use the SSIM metric evaluated on the MNIST test set. All models were trained for 3280 epochs. Note $(K,M)$ values are set to $(8,8)$ and $\beta=0.5$} \label{fig:reconstructions_mnist}
\end{figure}

Finally, we have also included the SSIM quantitative evaluation metric \cite{wang2004image} used to compare visual quality between two images on all pairs of original and reconstructed images from the test set. Qualitative and quantitative comparison of the reconstructions of data from MNIST dataset is in Figure \ref{fig:reconstructions_mnist}.
\begin{figure}[tb]
	\centering
	\includegraphics[height=6.0cm,width=8.0cm]{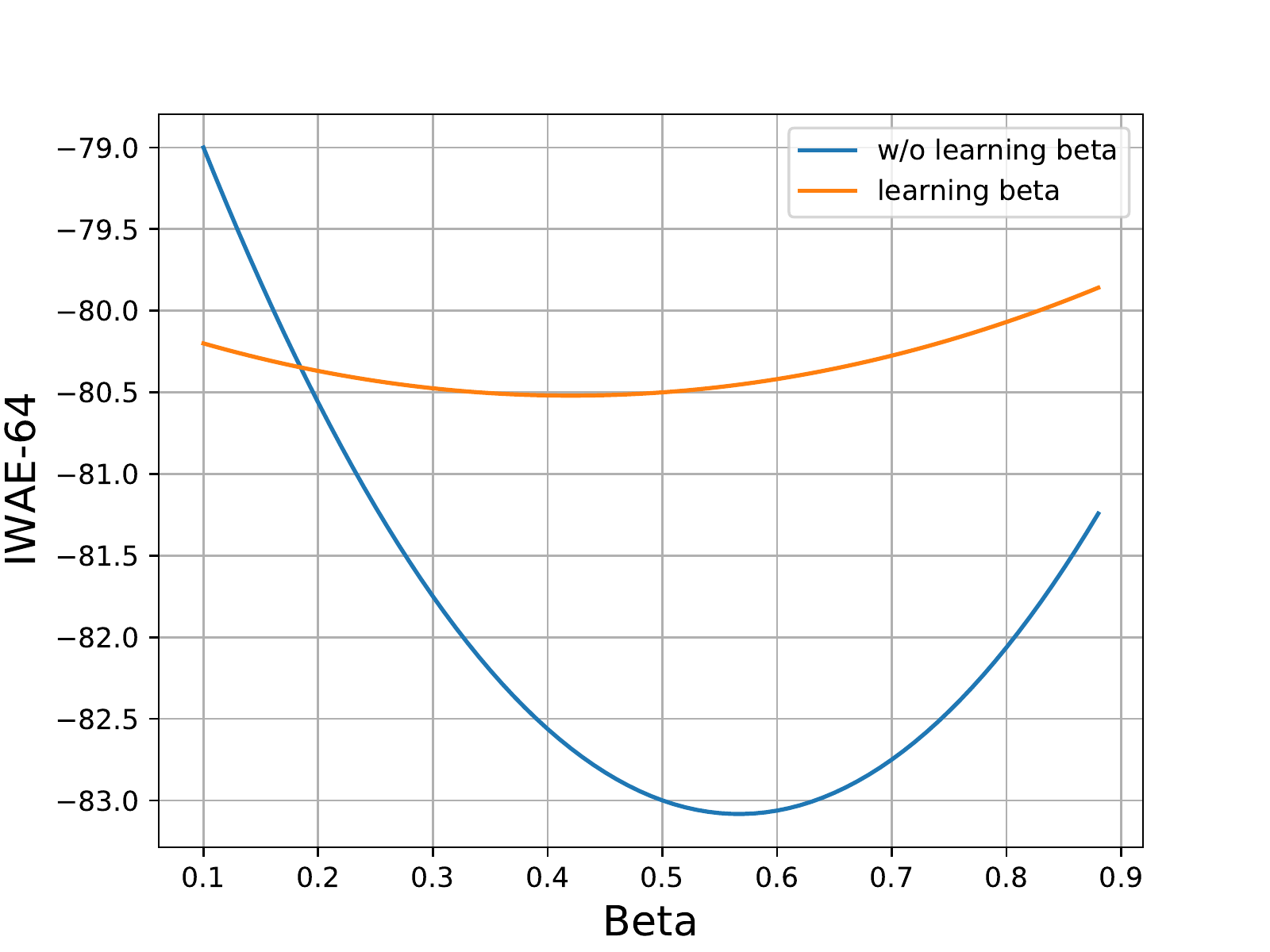}
	\caption{Illustration of IWAE-64 score with and without learning $\beta$. The average score when learning $\beta$ is -80.34\textpm0.49, whereas the average score is -81.99\textpm2.98 without learning $\beta$. }
	\label{fig:Beta_Learning}
\end{figure}

\begin{figure}[tb]
\begin{tcbitemize}[
    raster columns=6,
    raster halign=center,
    raster every box/.style={blankest},
    fonttitle=\small\bfseries 
    ]
\mysubfig{original \\ input}{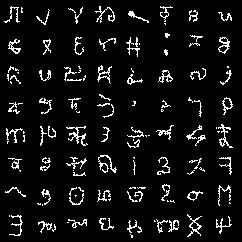}
\mysubfig{IWAE \\ 0.724$\pm$0.086}{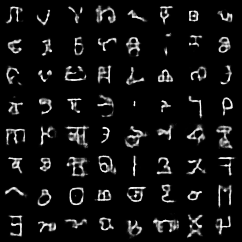}
\mysubfig{PIWAE \\ \textbf{0.743$\pm$0.082}}{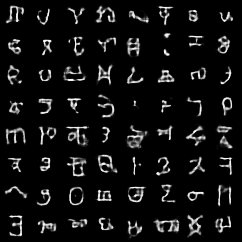}
\mysubfig{MIWAE \\ \textbf{0.745$\pm$0.082}}{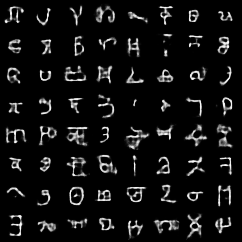}
\mysubfig{CIWAE \\ 0.719$\pm$0.098}{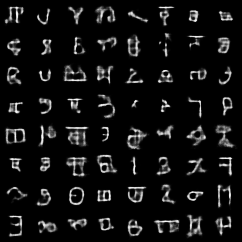}
\mysubfig{VAE \\ 0.717$\pm$0.099}{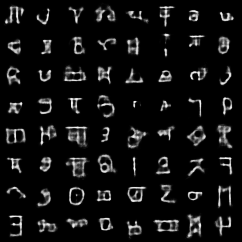}
\end{tcbitemize}

\caption{Qualitative and quantitative analysis of the reconstructed images on the Omniglot dataset with the referential model size. We use the SSIM metric evaluated on the Omniglot test set. All models were trained for 1000 epochs. Note $(K,M)$ values are set to $(8,8)$ and $\beta=0.5$} \label{fig:reconstructions_omni}
\end{figure}

\begin{figure}[tb]
\begin{tcbitemize}[
    raster columns=4,
    raster halign=center,
    raster valign=top,
    raster every box/.style={blankest},
    fonttitle=\small\bfseries 
    ]
\mysubfig{input \\ MNIST}{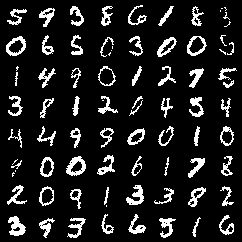}
\mysubfig{reconstruction by a Omniglot trained model}{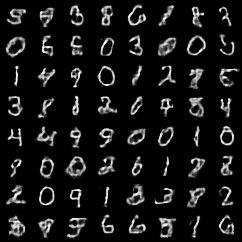}
\mysubfig{input \\ Omniglot}{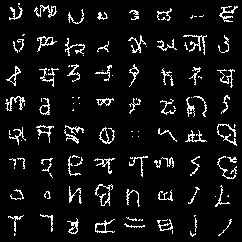}
\mysubfig{reconstruction by a MNIST trained model}{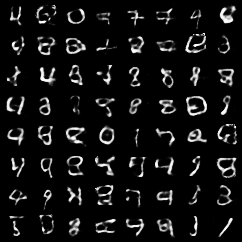}
\end{tcbitemize}

\caption{Generalization ability measured by reconstruction quality between models trained on one dataset and evaluated on another dataset. In all cases we used the MIWAE$_{(8, 8)}$ model for reconstruction. } \label{fig:reconstructions_cross}
\end{figure}

\subsection{Experiment on model generalizability}
Furthermore as a proxy measure for the generalizability of the trained models, we chose to evaluate models trained on one dataset and measure their SSIM score on another dataset. We note that while this experiment asks the models to perform a task which they were not trained on, due to the shared domain of datasets such as MNIST and Omniglot (handwritten digits and characters), we expect some overlapped knowledge in the  representational power of the models. This can be used in zero-shot learning scenarios and as an evaluation of the expected quality of reused weights between two data domains which is commonly used when fine-tuning models pre-trained on the ImageNet dataset. We show the quantitative evaluation of these results in Table \ref{fig:generalization} and show reconstructions in \ref{fig:reconstructions_cross}.

\subsection{Experiment on learning $\beta$ in CIWAE}
The purpose of this experiment is to measure how learning beta in CIWAE affects system performance. Our main intuition behind this experiment is that the optimal value of $\beta$ can be
different depending on the dataset, and finding this value can be a cumbersome task. To see if different $\beta$ leads to different result even in the same dataset, we first perform an experiment with different $\beta$. The following experiment is with learning $\beta$, to compare the performance with first experiment. Figure \ref{fig:Beta_Learning} shows the performance of the network for these two experiments in IWAE-64 metric. In the first experiment, when $\beta$ is set differently, the performance of the network is estimated to have -81.29\textpm 2.98, indicating that different $\beta$ can lead to a different result that can be suboptimal. On the other hand, when learning $\beta$, the average IWAE-64 metric is -80.34\textpm 0.49, suggesting that learning $\beta$ helps minimize the deviation caused by the different initial value of $\beta$. Considering that learning $\beta$ only adds one more optimization step using gradient descent, the gain by learning $\beta$ for alleviating the parameter tuning step would be advantageous.

\begin{table}[]
\caption{Generalization ability of different models variants. Note that all models are being evaluated by SSIM on the Omniglot datasets, while only the second row was trained on it.}
\label{fig:generalization}
\begin{tabular}{|l|lllll|}
\hline
Train:   & IWAE                         & PIWAE$_{(8, 8)}$                 & MIWAE$_{(8, 8)}$                 & CIWAE$_{\beta = 0.5}$             & VAE                          \\ \hline
MNIST   & 0.58$\pm$0.11 & \textbf{0.60$\pm$0.11}      & 0.59$\pm$0.12      & 0.58$\pm$0.12          & 0.57$\pm$0.12 \\
Omniglot & 0.72$\pm$0.09 & 0.74$\pm$0.08                & \textbf{0.75$\pm$0.08}      & 0.72$\pm$0.10          & 0.72$\pm$0.10 \\ \hline
\end{tabular}
\end{table}

\section{Conclusion}
\label{sec:conclusion}

In conclusion, we have reproduced the main methods proposed in \cite{rainforth2018tighter} as an extension to the model of \cite{burda2015importance}. Furthermore, we have extended the evaluation of these methods over a new dataset (and tested the generalization ability of these models) and with altered basic model architecture (approximately doubling the number of parameters). Reproduction of the methods from \cite{rainforth2018tighter} has yielded similar, but not exactly the same results, which might be due to implementation details or smaller scale of our experiments (namely only one run of the full 3280 epochs for each of the methods). Exploration of other hyper parameters, such as the choice of the architecture (dimensionality of the hidden and latent layers) has revealed larger performance gains than those achieved by varying the used learning method. We however note that these findings are orthogonal to each other and in practice we would strive to optimize both aspects - the size of the model architecture as well as the used method.

We also show that the $\beta$ hyper parameter of CIWAE can be learned, and more importantly that this approach leads to higher robustness in terms of independence on the initialization of this parameter. 

Evaluation of the reconstruction quality using the SSIM metric has indicated that the performance of the different models under this metric is comparable.

Lastly, our motivation to reproduce the methods of \cite{rainforth2018tighter} was to confirm their findings on other datasets. To this effect we run experiments with the Omniglot dataset, where we could see similar behaviour as the one seen on the reproduced and referential results on the MNIST dataset. If we had more time and computational resources for this assignment, we would try to extend it to other real world scenario datasets (as can be also seen in the development branch of our Github repository). Additionally, we wanted to provide and open-source implementation of these proposed methods, as there is no official code repository for this paper. This has made our efforts slightly more complicated, namely due to the fact that small implementation details can lead to considerable differences in the training progress.

{\small
\bibliographystyle{ref_style}
\bibliography{ref}

\begin{thebibliography}{10}\itemsep=-1pt

\bibitem{ba2014multiple}
Jimmy Ba, Volodymyr Mnih, and Koray Kavukcuoglu.
\newblock Multiple object recognition with visual attention.
\newblock {\em arXiv preprint arXiv:1412.7755}, 2014.

\bibitem{bornschein2014reweighted}
J{\"o}rg Bornschein and Yoshua Bengio.
\newblock Reweighted wake-sleep.
\newblock {\em arXiv preprint arXiv:1406.2751}, 2014.

\bibitem{burda2015importance}
Yuri Burda, Roger Grosse, and Ruslan Salakhutdinov.
\newblock Importance weighted autoencoders.
\newblock {\em arXiv preprint arXiv:1509.00519}, 2015.

\bibitem{cremer2017reinterpreting}
C. {Cremer}, Q. {Morris}, and D. {Duvenaud}.
\newblock {Reinterpreting Importance-Weighted Autoencoders}.
\newblock {\em Workshop at the International Conference on Learning
  Representations}, 2017.

\bibitem{dayan1995helmholtz}
Peter Dayan, Geoffrey~E Hinton, Radford~M Neal, and Richard~S Zemel.
\newblock The helmholtz machine.
\newblock {\em Neural computation}, 7(5):889--904, 1995.

\bibitem{gogate2007studies}
Vibhav Gogate, Bozhena Bidyuk, and Rina Dechter.
\newblock Studies in lower bounding probability of evidence using the markov
  inequality.
\newblock 2007.

\bibitem{gregor2014deep}
Karol Gregor, Ivo Danihelka, Andriy Mnih, Charles Blundell, and Daan Wierstra.
\newblock Deep autoregressive networks.
\newblock In {\em International Conference on Machine Learning}, pages
  1242--1250. PMLR, 2014.

\bibitem{kingma2015variational}
Diederik~P Kingma, Tim Salimans, and Max Welling.
\newblock Variational dropout and the local reparameterization trick.
\newblock {\em arXiv preprint arXiv:1506.02557}, 2015.

\bibitem{kingma2013auto}
Diederik~P Kingma and Max Welling.
\newblock Auto-encoding variational bayes.
\newblock {\em arXiv preprint arXiv:1312.6114}, 2013.

\bibitem{mnih2014neural}
Andriy Mnih and Karol Gregor.
\newblock Neural variational inference and learning in belief networks.
\newblock In {\em International Conference on Machine Learning}, pages
  1791--1799. PMLR, 2014.

\bibitem{neal1990learning}
Radford~M Neal.
\newblock Learning stochastic feedforward networks.
\newblock {\em Department of Computer Science, University of Toronto},
  64(1283):1577, 1990.

\bibitem{neal1992connectionist}
Radford~M Neal.
\newblock Connectionist learning of belief networks.
\newblock {\em Artificial intelligence}, 56(1):71--113, 1992.

\bibitem{nowozin2018debiasing}
Sebastian Nowozin.
\newblock Debiasing evidence approximations: On importance-weighted
  autoencoders and jackknife variational inference.
\newblock In {\em ICLR 2018 Conference}, February 2018.

\bibitem{rainforth2018tighter}
Tom Rainforth, Adam Kosiorek, Tuan~Anh Le, Chris Maddison, Maximilian Igl,
  Frank Wood, and Yee~Whye Teh.
\newblock Tighter variational bounds are not necessarily better.
\newblock In {\em International Conference on Machine Learning}, pages
  4277--4285. PMLR, 2018.

\bibitem{rezende2014stochastic}
Danilo~Jimenez Rezende, Shakir Mohamed, and Daan Wierstra.
\newblock Stochastic backpropagation and approximate inference in deep
  generative models.
\newblock In {\em International conference on machine learning}, pages
  1278--1286. PMLR, 2014.

\bibitem{backprop1986}
D.~E. Rumelhart, G.~E. Hinton, and R.~J. Williams.
\newblock {\em Learning Internal Representations by Error Propagation}, page
  318–362.
\newblock MIT Press, Cambridge, MA, USA, 1986.

\bibitem{salakhutdinov2010efficient}
Ruslan Salakhutdinov and Hugo Larochelle.
\newblock Efficient learning of deep boltzmann machines.
\newblock In {\em Proceedings of the thirteenth international conference on
  artificial intelligence and statistics}, pages 693--700. JMLR Workshop and
  Conference Proceedings, 2010.

\bibitem{salakhutdinov2008quantitative}
Ruslan Salakhutdinov and Iain Murray.
\newblock On the quantitative analysis of deep belief networks.
\newblock In {\em Proceedings of the 25th international conference on Machine
  learning}, pages 872--879, 2008.

\bibitem{smolensky1986information}
Paul Smolensky.
\newblock Information processing in dynamical systems: Foundations of harmony
  theory.
\newblock Technical report, Colorado Univ at Boulder Dept of Computer Science,
  1986.

\bibitem{wang2004image}
Zhou Wang, Alan~C Bovik, Hamid~R Sheikh, and Eero~P Simoncelli.
\newblock Image quality assessment: from error visibility to structural
  similarity.
\newblock {\em IEEE transactions on image processing}, 13(4):600--612, 2004.

\bibitem{williams1992simple}
Ronald~J Williams.
\newblock Simple statistical gradient-following algorithms for connectionist
  reinforcement learning.
\newblock {\em Machine learning}, 8(3-4):229--256, 1992.

\end{thebibliography}
}

\end{document}